\documentclass[letterpaper, 10 pt, conference]{ieeeconf}  

\IEEEoverridecommandlockouts                              

\usepackage{graphics} 
\usepackage{graphicx} 
\usepackage{amsmath} 
\usepackage{amssymb}  
\usepackage{amsthm}
\usepackage{bbm}
\usepackage{caption}
\usepackage{gensymb}
\usepackage{hyperref}
\usepackage{subcaption}
\usepackage[per-mode=symbol]{siunitx}
\usepackage{booktabs}
\usepackage[backend=biber,
    style=ieee,
    bibstyle=ieee,
    sortcites=true,   
    minbibnames=10,
    maxbibnames=20,
    mincitenames=1,
    maxcitenames=2,
    giveninits=true,
    uniquelist=false,
    backref=false]{biblatex}

\usepackage[dvipsnames,table,xcdraw]{xcolor}
\usepackage{makecell}
\usepackage{multirow}
\usepackage{siunitx}

\newcommand{\stepperiod}{T_{\text{s}}}                  

\newcommand{\CoMPos}{\boldsymbol{r}}                    
\newcommand{\CoMPosX}{r_x}                              
\newcommand{\CoMPosY}{r_y}                              

\newcommand{\CoMVel}{\dot{\boldsymbol{r}}}              
\newcommand{\CoMVelX}{\dot{r}_x}                        
\newcommand{\CoMVelY}{\dot{r}_y}                        
\newcommand{\dCoMVel}{\ddot{\boldsymbol{r}}}            

\newcommand{\footPos}{\boldsymbol{p}}                   
\newcommand{\footPosX}{p_x}                             
\newcommand{\footPosY}{p_y}                             

\newcommand{\dfootPos}{\hat{\boldsymbol{p}}}            
\newcommand{\dfootPosX}{\hat{p}_x}                      
\newcommand{\dfootPosY}{\hat{p}_y}                      

\newcommand{\ICP}{\boldsymbol{\xi}}                     
\newcommand{\ICPX}{\xi_x}                               
\newcommand{\ICPY}{\xi_y}                               
\newcommand{\dICP}{\dot{\boldsymbol{\xi}}}              

\newcommand{\velcmd}{\hat{\boldsymbol{v}}}              
\newcommand{\velcmdX}{\hat{v}_x}                        
\newcommand{\velcmdY}{\hat{v}_y}                        

\newcommand{\turningAngle}{\theta}                      
\newcommand{\contactSchedule}{\phi_{\text{contact}}}    
\newcommand{\phase}{\phi}                               

\newcommand{\remainingTs}{\delta T}                     

\newcommand{\thickhline}{\noalign{\hrule height 1.5pt}}

\newcommand{\ie}{i\/.\/e\/.,\/~}%
\usepackage[normalem]{ulem}  

\title{\LARGE \bf
Integrating Model-Based Footstep Planning with Model-Free Reinforcement Learning for Dynamic Legged Locomotion
}

\author{Ho Jae Lee$^{1}$, Seungwoo Hong$^{1}$, and Sangbae Kim$^{1}$
\thanks{$^{1}$All authors are with the Biomimetic Robotics Lab, Mechanical Engineering Department, Massachusetts Institute of Technology, Cambridge, Massachusetts,
            USA (e-mail: {\tt\small\{hjlee201, swhong, sangbae\}@mit.edu}). }%
\thanks{This work was supported by NAVER Labs.}
}

\begin{document}

\maketitle
\thispagestyle{empty}
\pagestyle{empty}

\begin{abstract}
In this work, we introduce a control framework that combines model-based footstep planning with Reinforcement Learning (RL), leveraging desired footstep patterns derived from the Linear Inverted Pendulum (LIP) dynamics. 
Utilizing the LIP model, our method forward predicts robot states and determines the desired foot placement given the velocity commands. 
We then train an RL policy to track the foot placements without following the full reference motions derived from the LIP model. 
This partial guidance from the physics model allows the RL policy to integrate the predictive capabilities of the physics-informed dynamics and the adaptability characteristics of the RL controller without overfitting the policy to the template model.
Our approach is validated on the MIT Humanoid, demonstrating that our policy can achieve stable yet dynamic locomotion for walking and turning. 
We further validate the adaptability and generalizability of our policy by extending the locomotion task to unseen, uneven terrain. 
During the hardware deployment, we have achieved forward walking speeds of up to 1.5 m/s on a treadmill and have successfully performed dynamic locomotion maneuvers such as 90-degree and 180-degree turns.

\end{abstract}

\section{INTRODUCTION} \label{sec:introduction}

Legged biological systems are capable of navigating through unstructured, complex, and discontinuous terrains, such as stepping stones. In the realm of legged robotics, researchers have long strived to enable legged robots to achieve mobility comparable to their natural counterparts, which would provide numerous practical real-world applications. 
However, designing controllers for legged robots is non-trivial because they have high degrees-of-freedom and their under-actuated floating base can only be indirectly controlled through external contact wrenches, making their equations of motion highly nonlinear and non-smooth.

\begin{figure}[tb]
  \centering
  \includegraphics[width=0.85\linewidth]{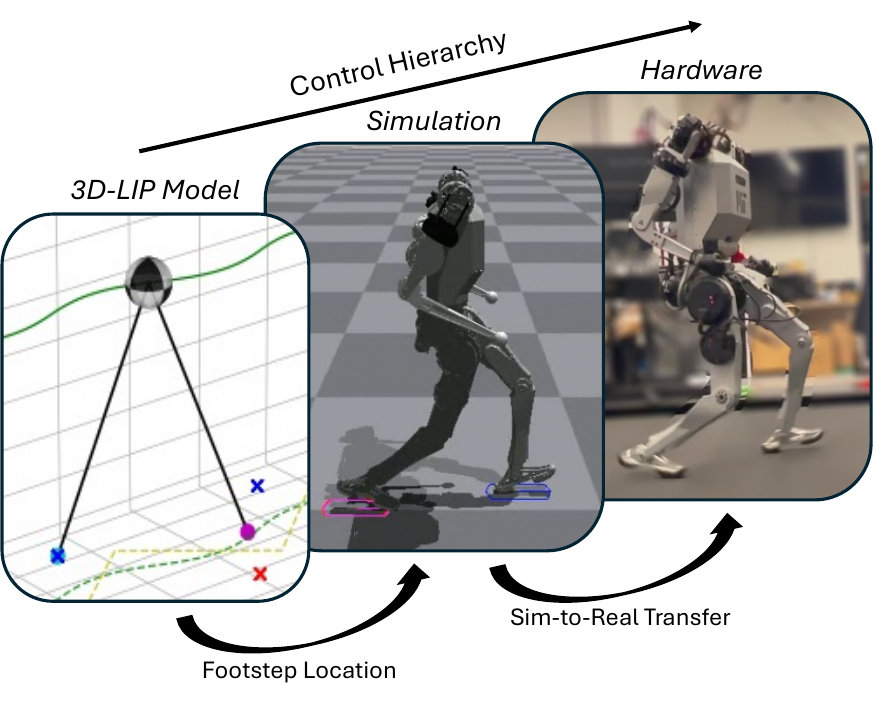}
  \caption{Our control hierarchy that employs a 3D-LIPM to determine the desired footstep location for locomotion. We train an RL policy to track the given steps and deploy the policy on MIT Humanoid. }
  \label{fig:overview}
\end{figure}

Model-based control approaches, such as reactive and predictive control methods, have emerged as a highly effective strategy for solving these complex control challenges, showcasing remarkable performance in quadrupedal and bipedal robotics applications
\cite{kajita20013d, hof2008extrapolated, englsberger2011bipedal,semini2016design,herdt2010online,rutschmann2012nonlinear,di2018dynamic,hong2022agile,kuindersma2016optimization}. 
The major advantage of the model-based control approach lies in leveraging insights from physics models to predict robot behavior, thereby enhancing controller design. 
In particular, foot placement emerges as one of the main components in model-based control on uneven or discontinuous terrain, providing an interface to control the robot through contact forces. 
Numerous studies have successfully used simplified models, such as the linear inverted pendulum model (LIPM)~\cite{kajita20013d, hof2008extrapolated, englsberger2011bipedal} to calculate foot placements or utilized optimization-based approaches that leverage simplified models like spring-loaded inverted pendulum (SLIP)~\cite{rutschmann2012nonlinear}, single rigid body dynamics (SRBD)~\cite{bledt2019implementing}, centroidal dynamics~\cite{grandia2023perceptive} for simultaneous computation of foot placements and contact forces.
However, these model-based control strategies that purely rely on simplified dynamics are inherently constrained by their simplifications and model mismatches, resulting in conservative locomotion that does not fully exploit the robot’s capabilities.

Parallel to these developments, model-free Reinforcement Learning (RL) has emerged as a powerful tool for robotic control, demonstrating remarkable success in managing complex, dynamic environments~\cite{kober2013reinforcement}. The application of model-free RL to both quadrupedal and bipedal robots has showcased a great performance in the given task and robustness against external perturbations~\cite{lee2020learning,miki2022learning,duan2022learning,xie2020learning}.
Nonetheless, the model-free RL lacks interpretability, making the process of reward shaping and hyperparameter tuning less straightforward and challenging. Furthermore, it has difficulty in generalizing learned policies to new tasks or environments without undergoing retraining. 
In response, numerous studies used heuristic-based references or model-based physics insights to inform and guide policy learning.
Specifically, some of the studies have employed heuristic or sampling-based methods to generate footstep locations and track these references using RL policy~\cite{duan2022learning, duan2022sim}. However, the absence of physical reasoning in footstep selection makes it challenging to accurately track target footstep locations while simultaneously maintaining balance without leading to instability or falls.
Other studies have used reduced-order models to guide the RL policy to follow the reference trajectories generated by these models~\cite{green2021learning, batke2022optimizing, jenelten2024dtc}. 
However, directly tracking the reference body and joint trajectories~\cite{jenelten2024dtc} or imitating the offline motion library~\cite{green2021learning, batke2022optimizing} from simplified models causes the RL policy to become overly aligned with the model, restricting exploration during training. Consequently, the resulting policy may be excessively constrained by the simplified model, failing to fully utilize the potential of whole-body dynamics.

In this work, we aim to bridge the gap between these two paradigms, integrating the physics-driven insights of model-based approaches with the adaptive and robust characteristics of RL. Specifically, we propose a hierarchical control framework that employs physics-informed step placements, utilizing linear inverted pendulum (LIP) dynamics to generate target step patterns, while concurrently training an RL policy to ensure the robot adheres to these prescribed step placements. 
This partial guidance from the physics-based template model prevents the policy from being confined to the model and results in a stable control policy capable of dynamic locomotion tasks, such as fast walking and sharp turns. Furthermore, our approach exhibits the robustness and adaptability inherent to RL policies, extending its capability to navigate unseen, uneven terrains by dynamically adjusting desired steps during the swing phase. The effectiveness of our approach is demonstrated through simulations and hardware experiments on the MIT Humanoid robot~\cite{saloutos2023design}, showcasing its potential in advancing robotic locomotion in complex environments (see video\footnote{\href{https://youtu.be/Z0E9AKt6RFo}{Supplementary Video Link}} and code\footnote{\href{https://github.com/hojae-io/ModelBasedFootstepPlanning-IROS2024.git}{Open-Source Code Link}}).

\section{BACKGROUND}

A single inverted pendulum that connects the supporting foot with its center of mass (CoM) via a massless telescopic leg is commonly used as a simplified model to represent bipedal locomotion \cite{kajita20013d}. 
By applying constraints to the inverted pendulum's motion, including a constant CoM velocity along the z-axis and a point-foot model without an actuated ankle joint, an analytical solution for the 3D-LIPM (Fig.~\ref{fig:lip_3d})~\cite{kajita20013d} can be formulated, governed by a linearly independent equation of motion:

\begin{equation} \label{eq:lip}
\dCoMVel = \omega_{0}^2(\CoMPos -  \footPos)
\end{equation}

where $\CoMPos = (\CoMPosX,\CoMPosY)^T$ denotes the position of the CoM, $\omega_{0} = \sqrt{g/z_{0}}$ indicates the natural frequency of the pendulum, and $\footPos = (\footPosX,\footPosY)^T$ represents the position of the foot. During the derivation of (\ref{eq:lip}), it is assumed that the foot is in contact with the ground. 

Integrating (\ref{eq:lip}) yields the "orbital energy"~\cite{kajita20013d}, leading to the formulation of the Instantaneous Capture Point (ICP), which is a point on the ground that the system comes to a stop if it were to instantaneously place its foot there~\cite{koolen2012capturability}:

\begin{equation} \label{eq:icp}
\ICP = \CoMPos + \frac{\CoMVel}{\omega_{0}}
\end{equation}
where $\ICP = (\ICPX,\ICPY)^T$ denotes the position of the ICP.
By differentiating (\ref{eq:icp}) with respect to time, and inserting (\ref{eq:lip}) into that equation, we obtain the ICP dynamics:

\begin{equation} \label{eq:icp_dynamics}
\dICP =\omega_{0}(\ICP - \footPos)
\end{equation}

We can derive the solution of (\ref{eq:icp_dynamics}) as follows:
\begin{equation} \label{eq:icp_traj}
\ICP(t) = e^{\omega_{0}t}\ICP(0) + (1 - e^{\omega_{0}t})\footPos
\end{equation}

These principles, described in (\ref{eq:icp_dynamics}) and (\ref{eq:icp_traj}), are pivotal for generating stable step patterns, as discussed in the next section.


\section{STEP PATTERN GENERATION ALGORITHMS}\label{sec:steppattern}
In this section, we describe the process of generating a suitable step pattern for the 3D-LIPM to achieve a velocity tracking task~\cite{hof2008extrapolated, englsberger2011bipedal}. We incorporate these strategies into the RL problem in Sec.~\ref{sec:rlproblem}, ensuring the bipedal robot aligns its foot placement with calculated step locations. 

\begin{figure}[tb]
    \centering
    \begin{subfigure}[b]{\linewidth}
        \includegraphics[width=\linewidth]{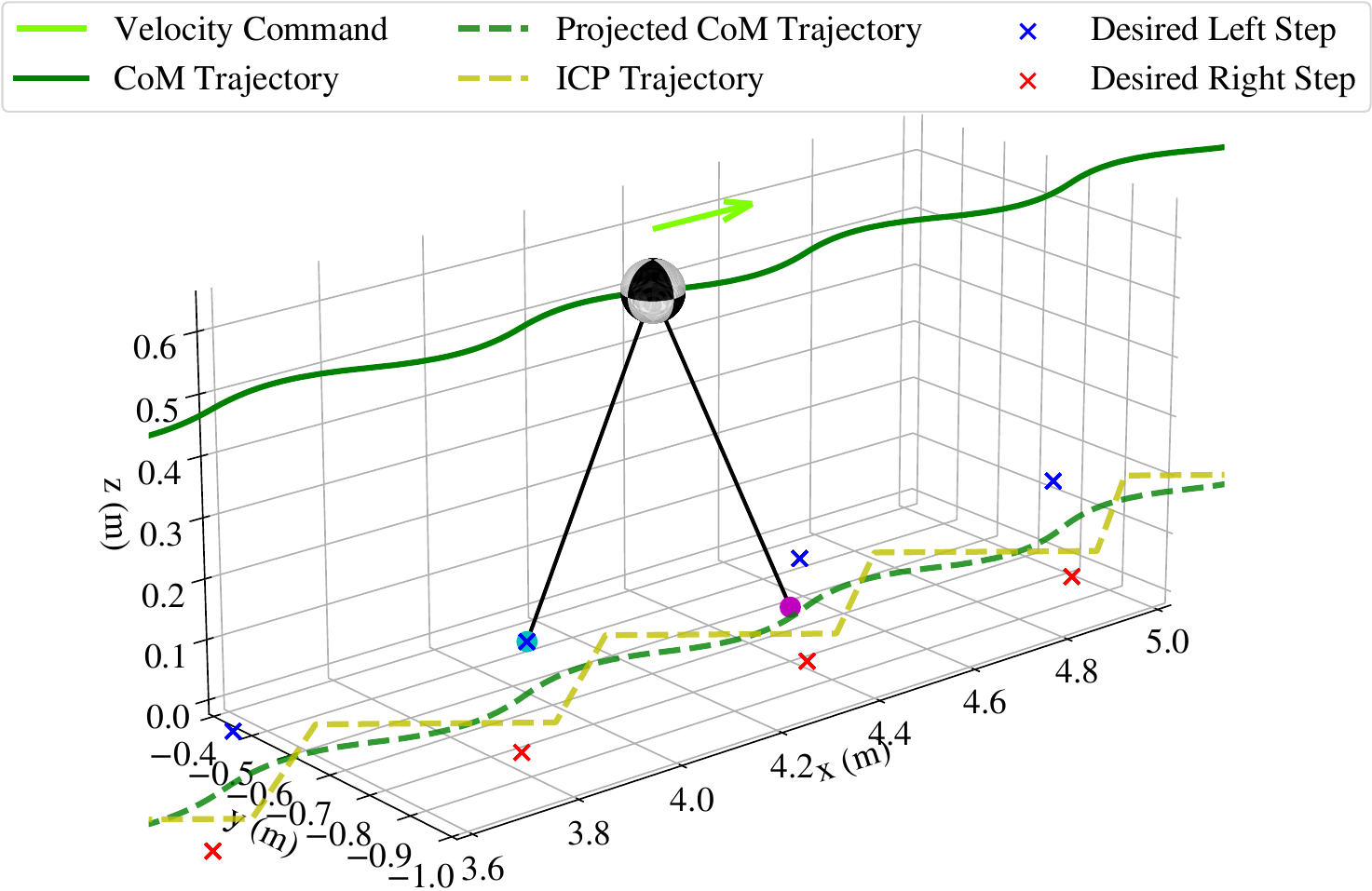}
        \caption{3D view}
        \label{fig:lip_3d}
    \end{subfigure}
    \begin{subfigure}[b]{\linewidth}
        \includegraphics[width=\linewidth]{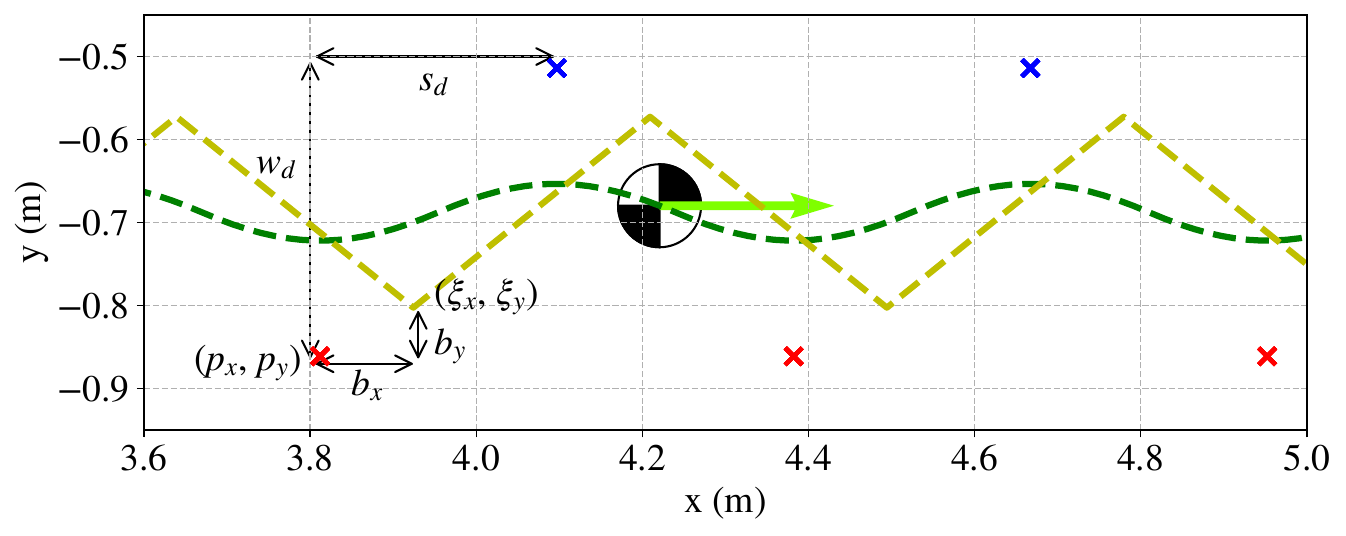}
        \caption{2D top-view}
        \label{fig:lip_2d}
    \end{subfigure}
    \caption{Step pattern generation algorithms for 3D-LIPM from 3D (Figure~\ref{fig:lip_3d}), 2D top-view (Figure~\ref{fig:lip_2d}) perspective. Figure~\ref{fig:lip_3d} depicts the LIPM with two legs. The LIP dynamics can predict the CoM trajectory (green lines, and green dashed lines). Our method calculates ICP trajectory (yellow lines) and adds offsets $(b_x, b_y)$ to the final ICP $(\ICPX^{\text{f}}, \ICPY^{\text{f}})$ to determine desired step locations for tracking velocity commands. Figure~\ref{fig:lip_2d} depicts the top view of the proposed method.}
    \label{fig:step_generation}
\end{figure}

\begin{figure*}[tb]
  \centering
  \includegraphics[width=0.9\textwidth]{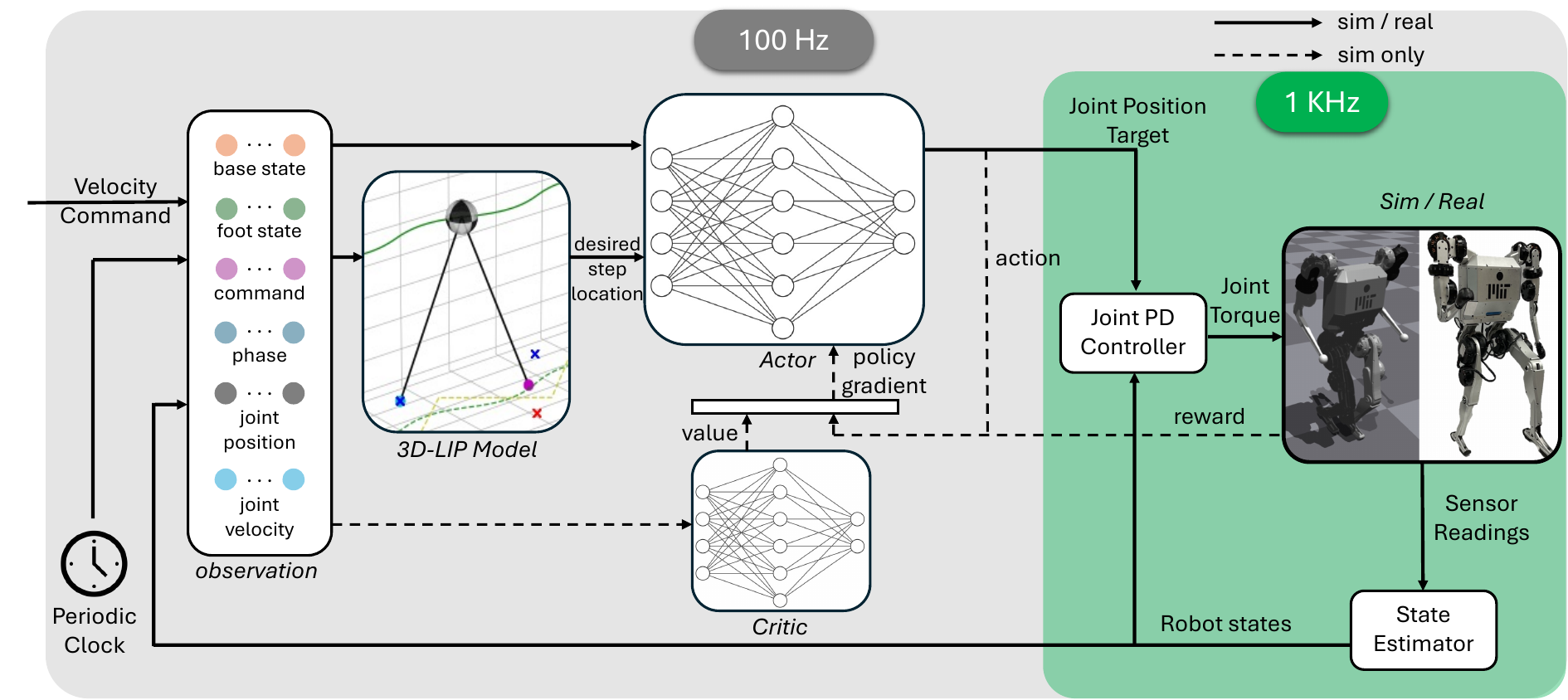}
  \caption{Overall control diagram and training framework for both learning in simulation and deployment to hardware. Step pattern generation algorithms generate the desired step location by utilizing the robot's CoM position, velocity, and foot states. These algorithms and NN update at a frequency of 100 Hz where both the actor and critic are trained using the PPO algorithm. Once the policy (actor) outputs joint position targets, the joint PD controller is evaluated at 1 KHz, and the command torques are sent to the motor. }
  \label{fig:controldiagram}
\end{figure*}

Our main objective is to track the desired base velocity command. Fig.~\ref{fig:lip_3d} outlines our step pattern generation algorithms, deriving the ICP trajectory and calculating the necessary offsets to determine the step locations for the left and right steps. Fig.~\ref{fig:lip_2d} presents a top-view of our algorithms.

We assume the LIPM moves along the positive $x$-axis. First, we calculate the desired step length $s_{d}$ based on the velocity command $\velcmd = (\velcmdX, \velcmdY)$:
\begin{equation} \label{eq:s_d}
s_{d}  =  |\velcmd|\cdot\remainingTs
\end{equation}
Here, $\delta T$ denotes the remaining step duration calculated by $\delta T = \stepperiod - t$, where $\stepperiod$ represents the user-defined step duration, and $t$ indicates the elapsed time since the beginning of the step. At the start of each step, $t$ is reset to 0 and progresses to $\stepperiod$. 

Similarly, we calculate the desired step width $w_{d}$ based on the step width command $\hat{w}$:
\begin{equation} \label{eq:w_d}
w_{d}  =  |\hat{w}|\cdot\remainingTs/\stepperiod
\end{equation}

Given the initial (\ie at the beginning of the step) body state $\CoMPos^{\text{o}} = (\CoMPosX^{\text{o}}, \CoMPosY^{\text{o}})$, $\CoMVel^{\text{o}} = (\CoMVelX^{\text{o}}, \CoMVelY^{\text{o}})$, we calculate the initial ICP $\ICP^{\text{o}} = (\ICPX^{\text{o}}, \ICPY^{\text{o}})$. Then we predict the LIP's final (\ie at the end of step) ICP $\ICP^{\text{f}} = (\ICPX^{\text{f}}, \ICPY^{\text{f}})$ after $\remainingTs$ using (\ref{eq:icp_traj}):  

\begin{equation}\label{eq:icp_final}
\begin{pmatrix}
\ICPX^{\text{f}} \\
\ICPY^{\text{f}} \\
\end{pmatrix}=
\begin{pmatrix}
\ICPX(\remainingTs) \\
\ICPY(\remainingTs) \\
\end{pmatrix}=
\begin{pmatrix}
e^{\omega_0\remainingTs}\cdot\ICPX^{\text{o}} + (1-e^{\omega_0\remainingTs})\cdot \footPosX \\
e^{\omega_0\remainingTs}\cdot\ICPY^{\text{o}} + (1-e^{\omega_0\remainingTs})\cdot \footPosY \\
\end{pmatrix}
\end{equation}

Based on Fig.~\ref{fig:lip_2d}, we observe that $s_{d}$ and $w_{d}$ can be readily expressed as follows:
\begin{equation}\label{eq:sdwd}
\begin{pmatrix}
    s_{d}  \\
    w_{d}  \\
\end{pmatrix} =
\begin{pmatrix}
    (\ICPX^{\text{f}} - \ICPX^{\text{0}})  \\
    (\ICPY^{\text{f}} - \ICPY^{\text{0}}) + 2(\ICPY^{\text{0}} - \footPosY)  \\
\end{pmatrix} 
\end{equation}

By inserting (\ref{eq:sdwd}) into (\ref{eq:icp_final}), we obtain the constant offset vector $(b_x, b_y)$, which when added to the final ICP, guarantees the step pattern has the desired step length $s_{d}$ and width $w_{d}$:

\begin{equation}\label{eq:bxby}
\begin{aligned}
\begin{pmatrix}
    b_x  \\
    b_y  \\
\end{pmatrix} &=
\begin{pmatrix}
    \ICPX^{\text{0}} - \footPosX  \\
    \ICPY^{\text{0}} - \footPosY  \\
\end{pmatrix} =
\begin{pmatrix}
    \frac{\ICPX^{\text{f}} - \ICPX^{\text{0}}}{e^{\omega_0\remainingTs}-1}  \\
    \frac{(\ICPY^{\text{f}} - \ICPY^{\text{0}}) + 2(\ICPY^{\text{0}} - \footPosY)}{e^{\omega_0\remainingTs}+1}  \\
\end{pmatrix} =
\begin{pmatrix}
    \frac{s_d}{e^{\omega_0\remainingTs}-1}  \\
    \frac{w_d}{e^{\omega_0\remainingTs}+1}  \\
\end{pmatrix}
\end{aligned}
\end{equation}

Then we determine the desired step location $\dfootPos=(\dfootPosX, \dfootPosY)^T$ by adding this constant offset vector to the final ICP~\cite{hof2008extrapolated}:
\begin{equation}\label{eq:pxpy_forward}
\begin{pmatrix}
    \dfootPosX  \\
    \dfootPosY  \\
\end{pmatrix} =
\begin{pmatrix}
    \ICPX^{\text{f}} - b_x  \\
    \ICPY^{\text{f}} + (-1)^nb_y)  \\
\end{pmatrix} 
\end{equation}
where $n$ indicates the step cycle (even $n$ for the left step and odd $n$ for the right step). 
In the case of turning in $xy$-plane, we modify (\ref{eq:pxpy_forward}) by rotating the constant offset vector:
\begin{equation}\label{eq:pxpy_rotation}
\begin{pmatrix}
    \dfootPosX \\
    \dfootPosY \\
\end{pmatrix} = 
\begin{pmatrix}
    \ICPX^{\text{f}} \\
    \ICPY^{\text{f}} \\
\end{pmatrix} +
\begin{pmatrix}
    cos(\turningAngle) & -sin(\turningAngle) \\
    sin(\turningAngle) & cos(\turningAngle) \\
\end{pmatrix}
\begin{pmatrix}
    -b_x \\
    (-1)^nb_y \\
\end{pmatrix}
\end{equation}
where $\turningAngle$ refers to the rotation angle around z-axis defined by $\turningAngle=tan^{-1}(\hat{v}_y/\hat{v}_x)$.
During the training, we calculate the desired step location only at the beginning of the step when $t = 0$ (\ie $\remainingTs=\stepperiod$). The detailed values for each user-defined variable are given in Table~\ref{tab:step_algorithms_command}.

\begin{table}[tb]
\centering
\caption{User-defined Variables for Step Pattern Generation Algorithms}
\label{tab:step_algorithms_command}
\begin{tabular}{l | c}
\hline
\textbf{Variable} & \textbf{Value} \\
\hline
Velocity commands $\velcmd$ & \( U^2[-2.0,~2.0] \) m/s\\
Step width command $\hat{w}$ & 0.3 m \\
Step duration $\stepperiod$ & 0.35 s \\
Base height $\hat{p}_{\text{base},z}$ & 0.62 m \\
Base heading $\hat{\theta}_{\text{base}}$ & $tan^{-1}(\hat{v}_y/\hat{v}_x)$ \\
\hline
\end{tabular}
\end{table}

\section{RL PROBLEM FORMULATION}\label{sec:rlproblem}
In this section, we describe our RL training framework to ensure the robot tracks the velocity commands and the step pattern generated by (\ref{eq:pxpy_rotation}).

Fig.~\ref{fig:controldiagram} shows the overview of our control framework.
Our control policy is a fully connected neural network with 3 hidden layers, each layer with 256 nodes. The policy takes as input the robot states, step commands derived from our proposed step pattern algorithms, and user velocity commands, and outputs the desired residual joint PD setpoints. We train our policy in the IsaacGym simulation engine using PPO \cite{schulman2017proximal} algorithms with parallelization of 4096 environments and input normalization. Detailed information on PPO hyperparameters can be found in Table~\ref{tab:hyperparameters}. Now, we introduce the state space, action space, and reward formulation for the RL problem.

\subsection{State Space}
\label{sec:statespace}
The state space of our policy consists of the observed robot states, step commands, and user-defined velocity commands with a size of $\mathcal{S}\in\mathbb{R}^{51}$. 
In detail, $\mathcal{S}$ includes the base height, base linear velocity in the world frame, base angular velocity, projection of the gravity vector in the base frame, left and right foot location and heading in the base frame, left and right desired step location and heading in the base frame, velocity commands, phase clock in sine and cosine functions, joint position, joint velocity. The phase identifiers indicate the swing and stance phase of each foot through the contact scheduler. The base states are measured through the phase-based state estimator~\cite{mit-biomimetics_cheetah-software_2023} that assumes the foot contact on the ground at the specified contact schedule. 

\subsection{Action Space}
\label{sec:actionspace}
We define the action space $\mathcal{A}\in\mathbb{R}^{10}$ as the desired residual joint PD setpoints $\Delta\hat{\boldsymbol{q}}$, representing a deviation from the nominal joint position $\boldsymbol{q}^\text{ref}$ for hip yaw, hip abduction, hip pitch, knee and ankle joint respectively. 
The action from our policy is updated at a frequency of 100 Hz and fed into the joint PD controller. Then, the fixed-gain joint PD controller operates at 1 kHz.
To be specific, the joint PD controller uses the following equation to convert the action into the desired torque command:
\begin{equation}
\hat{\boldsymbol{\tau}} = \mathbf{K_p}(\boldsymbol{q}^\text{ref} + \Delta\hat{\boldsymbol{q}} - \boldsymbol{q}) + \mathbf{K_d}(\mathbf{0} - \dot{\boldsymbol{q}})
\end{equation}
For the joint PD controller's gains, we have configured $\mathbf{K_p}$ to $diag(30, ..., 30)$, and $\mathbf{K_d}$ to $diag(1, ..., 1)$.

\subsection{Rewards}
\label{sec:rewards}

\begin{table}[tb]
\centering
\caption{PPO Hyperparameters}
\label{tab:hyperparameters}
\begin{tabular}{l | c}
\hline
\textbf{Parameter} & \textbf{Value} \\
\hline
Horizon (H) & 24 \\ 
Adam learning rate & $1 \times 10^{-5}$ \\
Number of epochs & 5 \\
Number of mini-batches & 4 \\
Discount ($\gamma$) & 0.99 \\
Clipping parameter ($\epsilon$) & 0.2 \\
Max gradient norm & 1 \\
\hline
\end{tabular}
\end{table}

\begin{table}[tb]
\renewcommand*{\arraystretch}{1.4}
    \centering
    \caption{Regularization Rewards}
    \begin{tabular}{|c|c|c| } 
     \hline
     \textbf{Reward} & \textbf{Weight} & \textbf{Expression} \\ 
     \thickhline
     Joint torques & 1e-4 & \(- \lvert \boldsymbol{\tau} \rvert^2\) \\ 
     \hline
     Joint torque limits & 1e-2 & \(-  \operatorname{max} (\lvert \boldsymbol{\tau} \rvert - 0.9\boldsymbol{\tau}_{max}, 0)\) \\ 
     \hline
     Joint velocity & 1e-3 & \(- \lvert \dot{\boldsymbol{q}} \rvert^2\) \\  
     \hline
     Joint limits & 10 & \(- \operatorname{clip} (\lvert \boldsymbol{q} \rvert - 0.9\boldsymbol{q}_{max}, 0, 1)\) \\ 
     \hline
     Action smoothness 1 & 1e-3 & \(- \lvert (\boldsymbol{a}_{t} - \boldsymbol{a}_{t-1})/\Delta t \rvert ^2\) \\ 
     \hline
     Action smoothness 2 & 1e-4 & \(- \lvert (\boldsymbol{a}_{t} - 2\boldsymbol{a}_{t-1} + \boldsymbol{a}_{t-2})/\Delta t \rvert ^2\)  \\ 
     \hline
     Hip joint regularization & 1.25 & $\exp(-(\boldsymbol{q}_{\text{hip},xz})^2/\sigma)$ \\  
     \hline
     Base roll-pitch velocity & 1e-2 & $-(\omega_{\text{base}, x}^2 + \omega_{\text{base}, y}^2)$ \\ 
     \hline
     Base z-axis velocity & 1e-1 & \(- \lvert v_{\text{base}, z} \rvert^2\) \\  
      \hline
     Base tilting & 1 & $\exp(-(g_{\text{base},x}^2 + g_{\text{base},y}^2)/\sigma)$ \\
     \hline
     Termination & 100 & 
     $\begin{cases}
        -1, \; \operatorname{self-collision}, \\
        -1, \; \lvert \boldsymbol{v}_{\text{base}} \rvert \geq 10 \text{ [m/s]}, \\  
        -1, \; \lvert \boldsymbol{\omega}_{\text{base}} \rvert \geq 5 \text{ [rad/s]}, \\
        -1, \; g_{\text{base},x}, g_{\text{base},y} \geq 0.7, \\
        -1, \; p_{\text{base},z} < 0.3 \text{ [m]}, \\
        0, \; \operatorname{otherwise}.
     \end{cases}$ \\
     \hline
     
     \end{tabular} \label{tab:rewards}
\end{table}

We formulate the reward structure to ensure the robot tracks the desired step location while maintaining stability and adaptability. Since the desired step location is derived based on the LIPM, we incorporate specific rewards to satisfy the assumption of the LIPM. To retain the inherent flexibility and adaptability characteristic of RL policy, however, we do not impose explicit rewards to follow the LIPM's CoM trajectory.

The overall reward function is formulated as follows:

\begin{equation}
r = r_{bh} + r_{bo} + r_{vt} + r_{cs} + r_{\text{Regularization}}
\end{equation}

First, to address the LIPM's assumption of a constant height, we introduce a reward $r_{bh}$ that encourages the robot to keep a constant base height $\hat{p}_{\text{base},z}$:

\begin{equation}
r_{bh} = \exp(-(\hat{p}_{\text{base},z} - p_{\text{base},z})^2/\sigma)
\end{equation}

Given that the LIPM is represented solely by a point mass and lacks any orientation, it offers no direct control over the robot's orientation. Therefore, we assume that the robot's base should consistently orient towards the desired base heading $\hat{\theta}_{\text{base}}$ direction. To encapsulate this concept, the reward $r_{bo}$ is designed:

\begin{equation}
r_{bo} = 2 \exp(-\lvert\hat{\theta}_{\text{base}} - \theta_{\text{base}}\rvert/\sigma)
\end{equation}

The desired step location calculated by step pattern generation algorithms in Sec.~\ref{sec:steppattern} results in the LIPM's passive dynamics naturally fulfilling the velocity command $\velcmd$.  Given the robot's deviation from the LIPM, we implement the velocity tracking reward $r_{vt}$ to ensure tracking of the velocity command $\velcmd$:

\begin{equation}
r_{vt} = 4 \exp(-(\frac{\velcmd - \boldsymbol{v}_{\text{world}}}{1 + \lvert\velcmd\rvert})^2/\sigma)
\end{equation}

Upon determining the desired step location, the robot must place its foot at this location for the specified step duration. The reward $r_{cs}$ is crafted to incentivize the robot to conform to the contact schedule at the desired step location:

\begin{equation} \label{eq:contact}
r_{cs} = 9 (\mathbbm{1}_{\text{r,contact}} - \mathbbm{1}_{\text{l,contact}}) \contactSchedule \cdot \exp(-(||\dfootPos - \footPos||_2)/\sigma)
\end{equation}

Here, $\mathbbm{1}_{\text{r,contact}}$ and $\mathbbm{1}_{\text{l,contact}}$ are indicator functions for right and left foot ground contact, respectively. The contact schedule $\contactSchedule$ is a continuous function that oscillates between -1 and 1 across each two-step duration $2\stepperiod$:

\begin{equation}
\contactSchedule = \frac{sin(2\pi\phase)}{\sqrt{sin^2(2\pi\phase) + 0.04}}~,~~ \phase = \frac{t'}{2\stepperiod}
\end{equation}
where $t'$ denotes the elapsed time from the start of the right-foot step, which is reset to 0 every two-step cycle, $2\stepperiod$.

Furthermore, to mitigate any undesirable motions, a set of regularization rewards $r_{\text{Regularization}}$ is imposed to penalize excessive joint torque, velocities, unnecessary angular motion, policy termination due to falls, etc (see Table~\ref{tab:rewards}). The reward shaping parameter $\sigma$ for the exponential function is set to 0.25 during training.

\section{EXPERIMENT RESULTS} \label{sec:results}

We now present our simulation and hardware test results on MIT Humanoid to evaluate the effectiveness of our approach. 
The training process takes about three hours of wall clock time using a Nvidia GeForce 3090 GPU. 

\subsection{Simulation Results}
\label{subsec:simulation}

\textit{1) Velocity tracking performance}: 
\begin{figure}[tb]
  \centering
  \includegraphics[width=\linewidth]{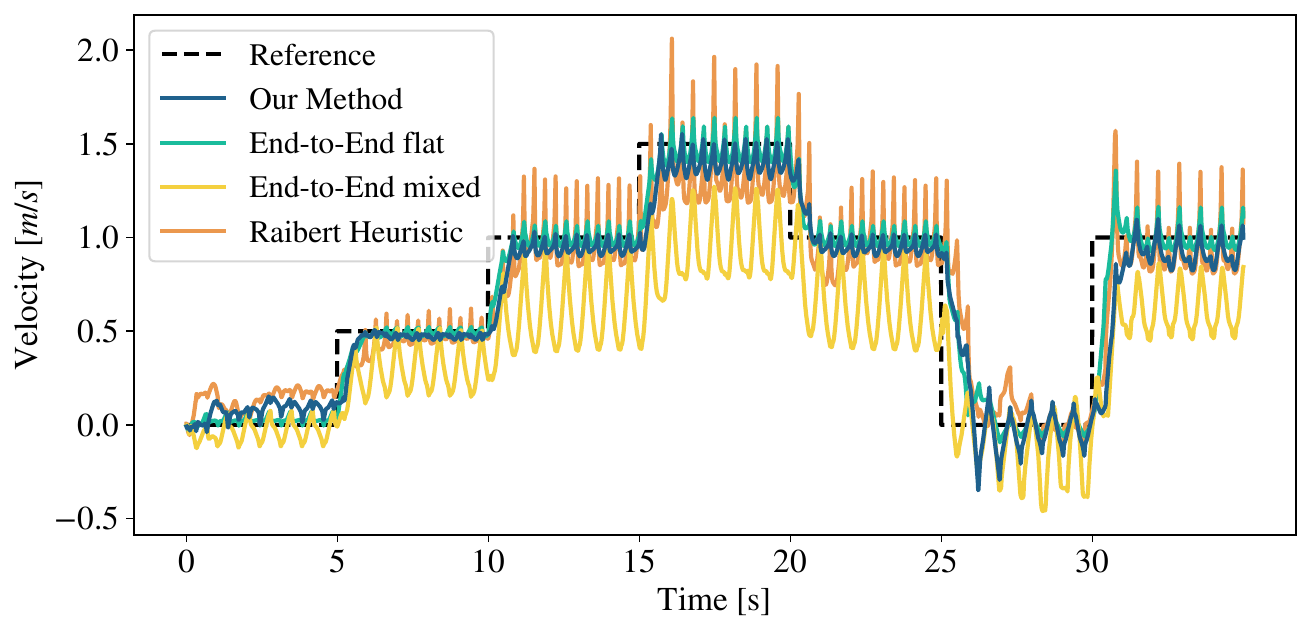}
  \caption{Comparison of velocity tracking performance between our method, End-to-End policies trained on flat terrain versus mixed terrains (flat, rough, and gap), and Raibert heuristic policy. Commands were given in flat terrain. Our method exceeds the performance of the End-to-End approach trained on varied terrains and shows comparable results to the End-to-End policy trained exclusively on flat terrain.}
  \label{fig:vx_comparison}
\end{figure}
Fig.~\ref{fig:vx_comparison} presents the velocity tracking performance of our method compared to: 1) End-to-End policy, which is trained to track the velocity commands without foot placement constraints; and 2) Raibert heuristic~\cite{raibert1983dynamically} policy, which replaces step pattern generation algorithms with Raibert heuristic. Our method and Raibert heuristic policy are trained exclusively on flat terrain. Additionally, we train two End-to-End policies: one on flat terrain; and the other on multiple terrains, including not only flat but also rough and gap-containing terrains. This plot depicts that our method exceeds the velocity tracking performance of the End-to-End policy trained across these diverse terrains. It also shows that our tracking accuracy is comparable to the End-to-End policy trained solely on flat terrain, which is recognized for its proficiency in single-task scenarios. The results validate that our method reliably tracks velocities up to 2.0 m/s. Furthermore, we observe that with the application of lateral velocity commands, the robot can execute dynamic maneuvers, including 90-degree and 180-degree turns.

\textit{2) Learning desired step duration}: 
\begin{figure}[tb]
    \centering
    \includegraphics[width=1.\linewidth]{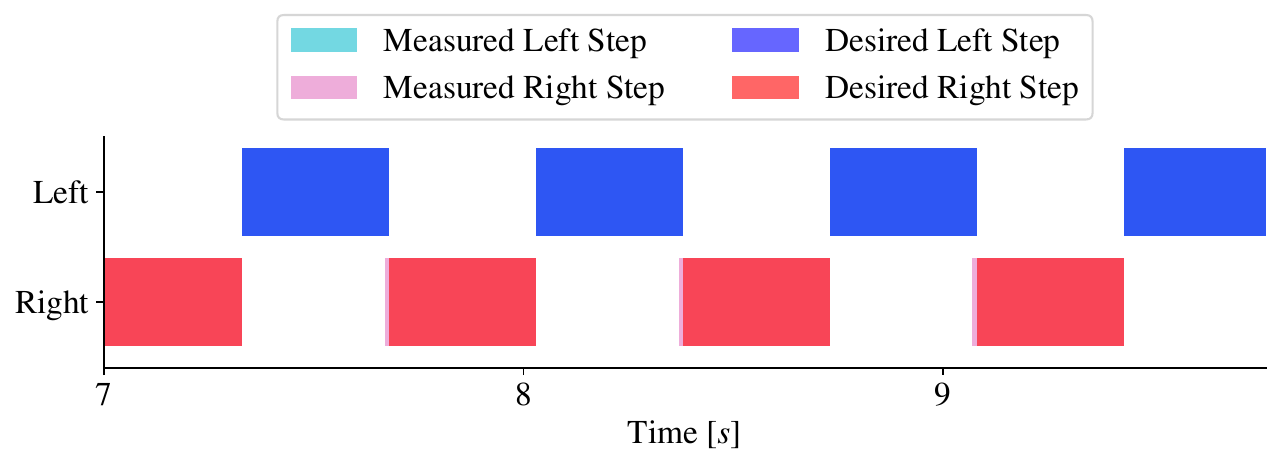}
    \caption{Learned foot contact schedule. The step duration $\stepperiod$ that was set to 0.35 seconds was encouraged by a contact schedule reward. The error between the measured and desired foot contact schedule is less than 0.01 seconds.}
    \label{fig:stepper_gait_sequence}
\end{figure}
Fig.~\ref{fig:stepper_gait_sequence} shows the step duration $\stepperiod$ learned by the policy. Throughout the training phase, $\stepperiod$ was fixed at 0.35 seconds, indicating the ground contact duration for a single step. In our setting, a foot is considered to be in contact with the ground if either the toe or heel is touching the ground. This behavior is encouraged through a contact schedule reward (\ref{eq:contact}). The plot confirms that the policy has successfully learned to maintain ground contact for 0.35 seconds for each leg, alternating between the left and right. This consistent step sequence is subsequently beneficial for employing a phase-based state estimator in hardware deployment.

\textit{3) Tracking desired step location}: 
\begin{figure}[tb]
    \centering
    \includegraphics[width=.85\linewidth]{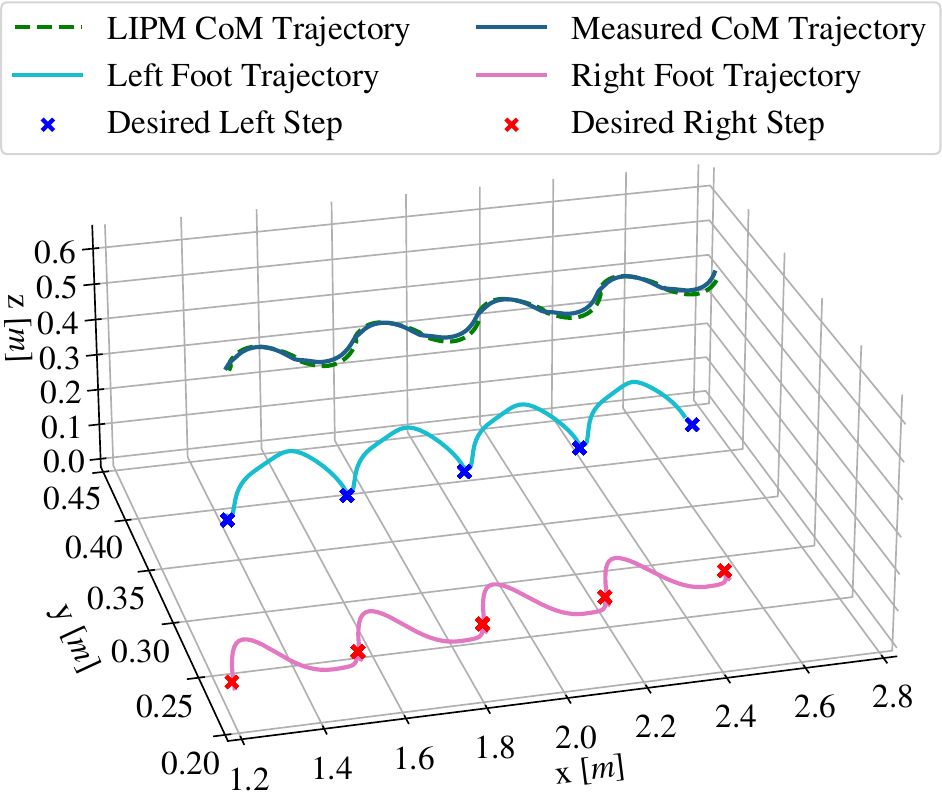}
    \caption{ Tracking desired step location and its resulting foot trajectory. Both left and right foot trajectories show a consistent and smooth path that culminates in an accurate touchdown at the target step locations. A notable observation is the robot's measured CoM trajectory exhibits a close correspondence to the LIPM CoM trajectory}
    \label{fig:stepper_foot_trajectory}
\end{figure}
Fig.~\ref{fig:stepper_foot_trajectory} shows the robot's successful tracking of desired step location generated by step pattern generation algorithms. Both right and left foot trajectories form a smooth and regular trajectory ensuring accurate touch down on the target step location.
Notably, the measured CoM trajectory of the robot closely aligns with the analytical LIPM CoM trajectory. 
This behavior is attributed to the implementation of the rewards that encourage the robot to satisfy the assumptions of LIPM.

\textit{4) Extension to rough terrain and gap terrain}:
\begin{figure*}[tb]
  \centering
  \begin{subfigure}[b]{0.48\textwidth}
    \includegraphics[width=\textwidth]{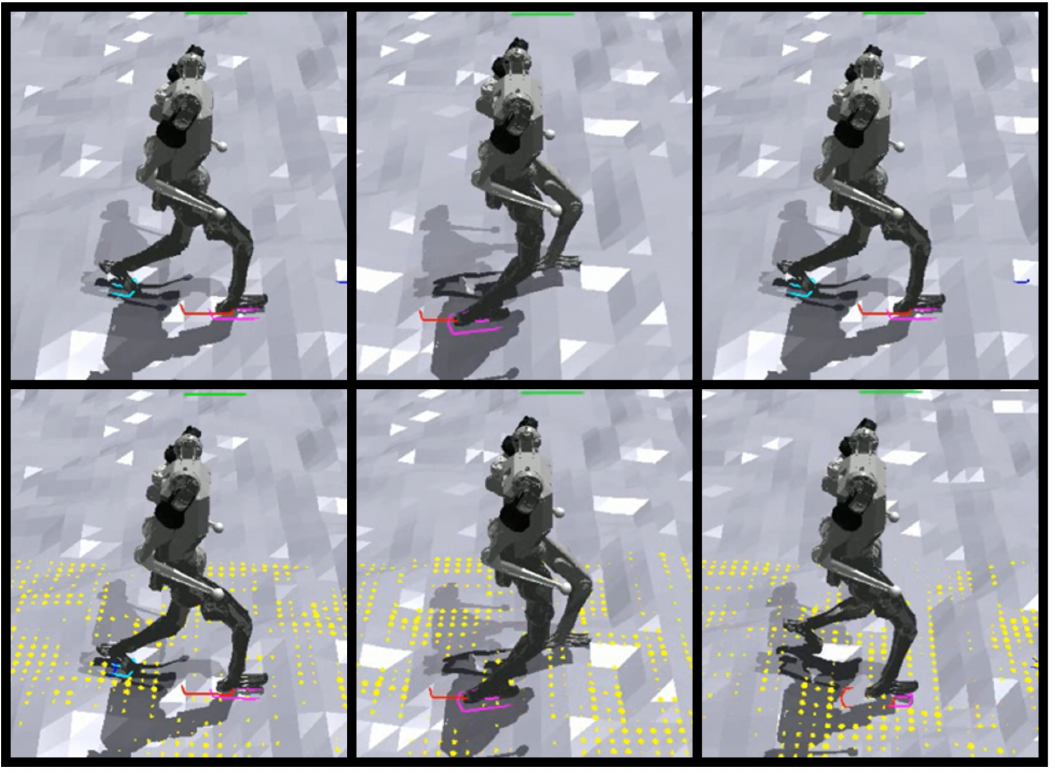}
    \caption{Rough terrain}
    \label{fig:rough}
  \end{subfigure}
  \hfill 
  \begin{subfigure}[b]{0.48\textwidth}
    \includegraphics[width=\textwidth]{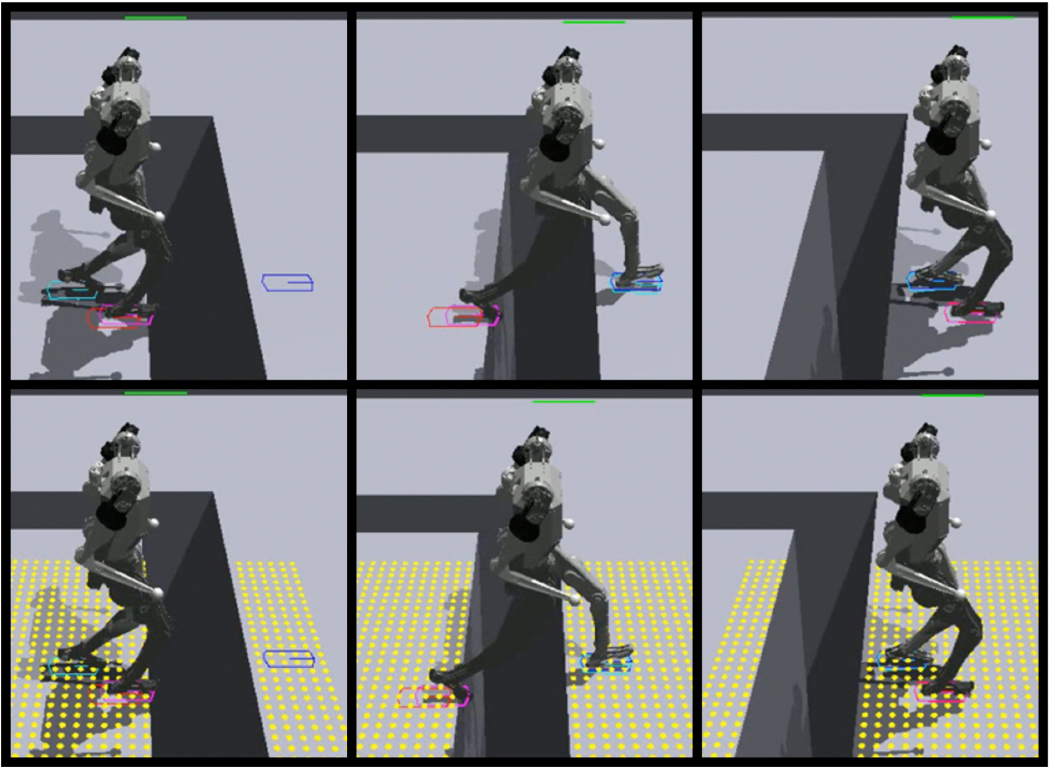}
    \caption{Gap terrain}
    \label{fig:gap}
  \end{subfigure}
  \caption{
  Adaptive locomotion on varied terrain. For rough terrain (Fig.~\ref{fig:rough}), the policy dynamically updates the desired step location by modifying $\remainingTs$ in the step pattern generation algorithms to compensate for the irregularities in terrain that affect the robot's CoM height. On gap terrain (Fig.~\ref{fig:gap}), the policy adjusts step location to the nearest flat surface, demonstrating the robot's capability to navigate discontinuities in the surface and maintain a forward velocity $\velcmdX$.
  }
  \label{fig:rough_gap}
\end{figure*}

\begin{figure}[tb]
    \centering
    \includegraphics[width=0.95\linewidth]{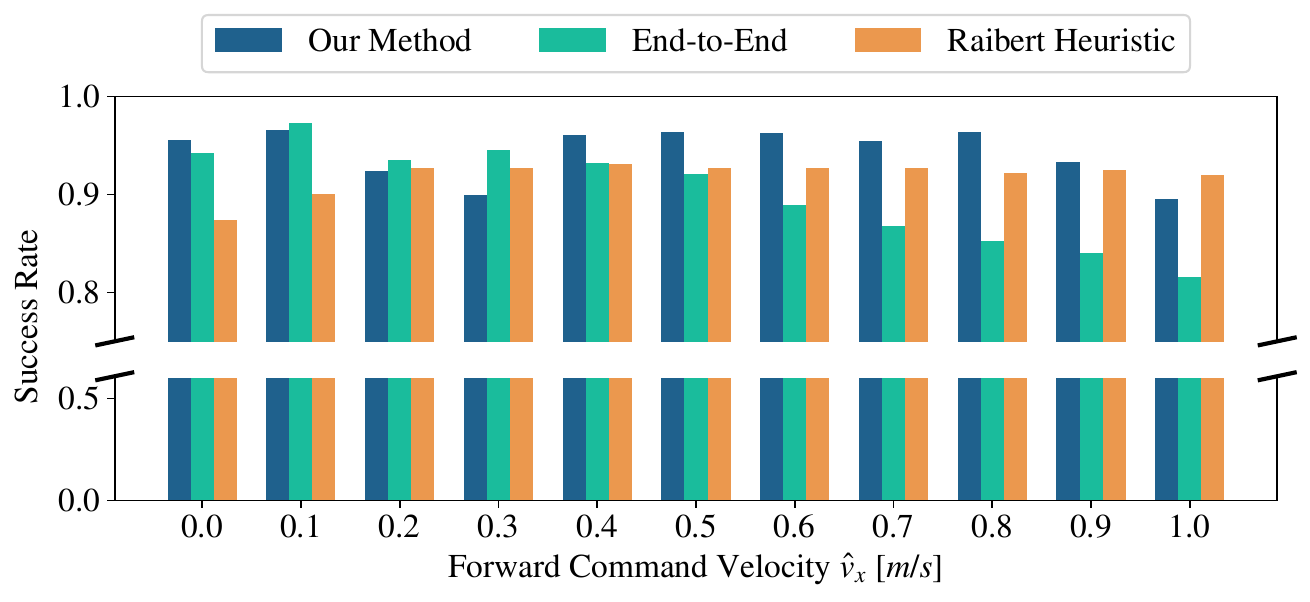}
    \caption{Success rate for walking on the rough terrain. To navigate rough terrain, we modify the robot's desired step locations at each time step to account for the deviation from the LIPM dynamics caused by uneven terrain. These adjustments, along with step elevation refinements based on heightmap data, helped maintain the robot's balance while tracking velocity command. Our method proved more effective than an End-to-End and Raibert heuristic policy trained on flat terrain, as evidenced by on average a higher success rate in maintaining forward velocity without falling.}
    \label{fig:Rough_success_rate}
\end{figure}

To evaluate the adaptability of our policy to unseen and uneven terrains, we conducted tests on both rough and gap terrain (Fig.~\ref{fig:rough_gap}). For the rough terrain, (Fig.~\ref{fig:rough}), we implemented dynamic adjustments of the desired step location by modifying $\remainingTs$ in equations (\ref{eq:s_d})-(\ref{eq:icp_final}) every time step. This approach compensates for the inevitable deviations from the LIP dynamics due to the rough terrain's impact on the constancy of the robot's CoM height. Additionally, we refined the desired step elevation in accordance with the ground height data obtained from a heightmap. The efficacy of our method was quantified by comparing it to an End-to-End policy and Raibert heuristic policy, originally trained on flat terrain, using a success metric defined by the robot's ability to maintain a predetermined forward velocity command $\velcmdX$ for five seconds without falling. As depicted in Fig.~\ref{fig:Rough_success_rate}, our policy showed on average a higher success rate.
In gap terrain scenarios (Fig.~\ref{fig:gap}), if the desired step location falls into a gap, we adjust it to the closest flat ground using heightmap data.
Through these deployments on both rough and gapped terrains, we have validated the robustness and adaptability of our policy: it can successfully modify the desired step location in response to real-time terrain alterations, thereby sustaining effective locomotion.

\subsection{Hardware Results}
\label{subsec:hardware}

We successfully transferred the policies developed in simulation to robot hardware, showcasing the robust sim-to-real transfer capabilities of our policies (Fig.~\ref{fig:forward_turning}). The robot demonstrated the ability to maintain a consistent height and precisely track the desired step locations for the given step duration $\stepperiod$. To compensate for state estimator noise, we dynamically modified the step locations at each timestep. The performance was evaluated through two specific locomotion tasks:


\begin{figure*}[tb]
  \centering
  \begin{subfigure}[b]{0.48\textwidth}
    \includegraphics[width=\textwidth]{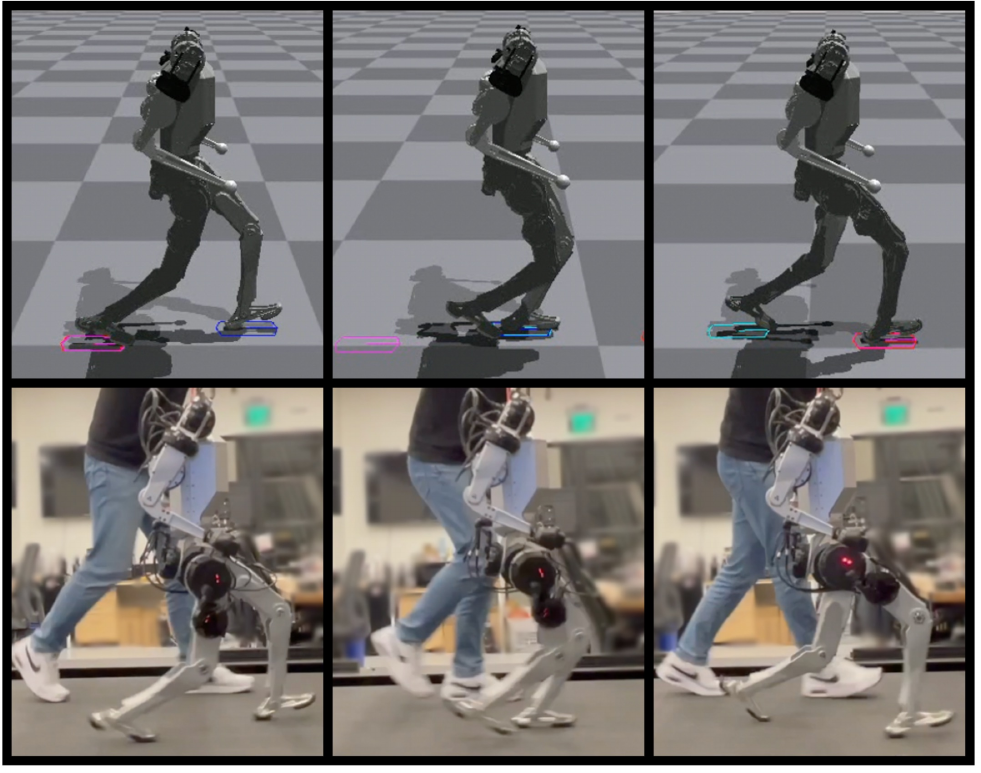}
    \caption{Forward Walking}
    \label{fig:forward_walking}
  \end{subfigure}
  \hfill 
  \begin{subfigure}[b]{0.48\textwidth}
    \includegraphics[width=\textwidth]{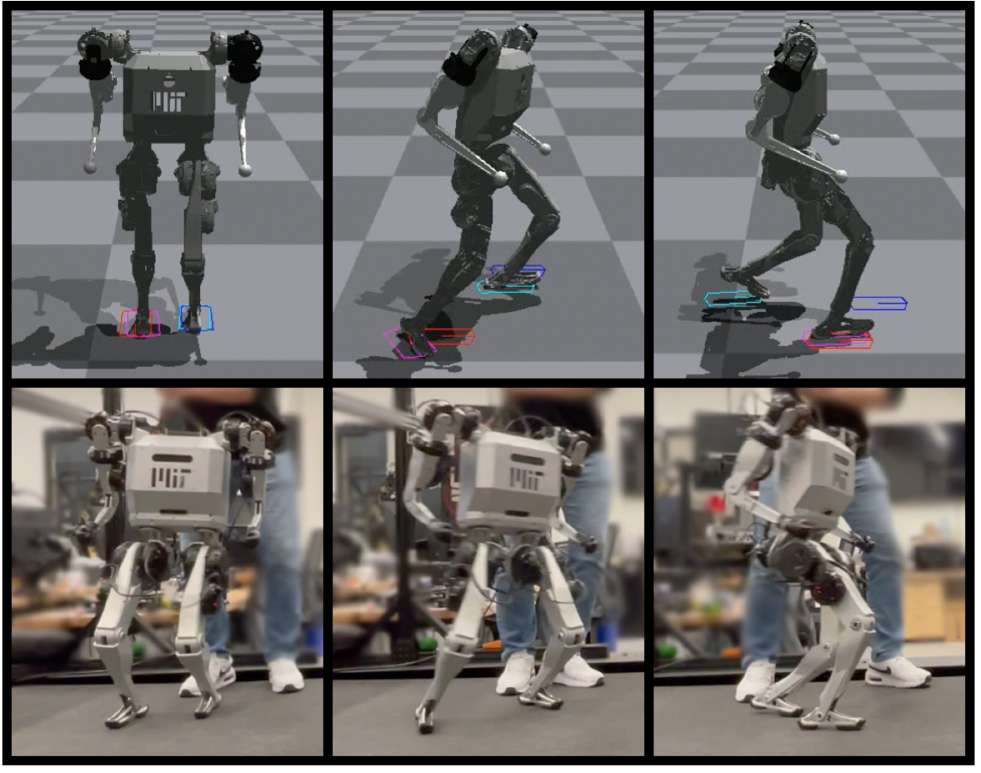}
    \caption{90\degree~Turning}
    \label{fig:turning}
  \end{subfigure}
  \caption{Hardware experiment results for forward walking and 90-degree turning. We successfully achieved forward walking at speeds up to 1.5 m/s and executed 90-degree turns, both featuring a heel-to-toe motion during touchdown similar to human walking. These motions in the hardware precisely mirrored those observed in the simulation environment.}
  \label{fig:forward_turning}
\end{figure*}

\begin{figure}[!ht]
  \centering
  \begin{subfigure}[b]{0.8\linewidth}
    \includegraphics[width=\linewidth]{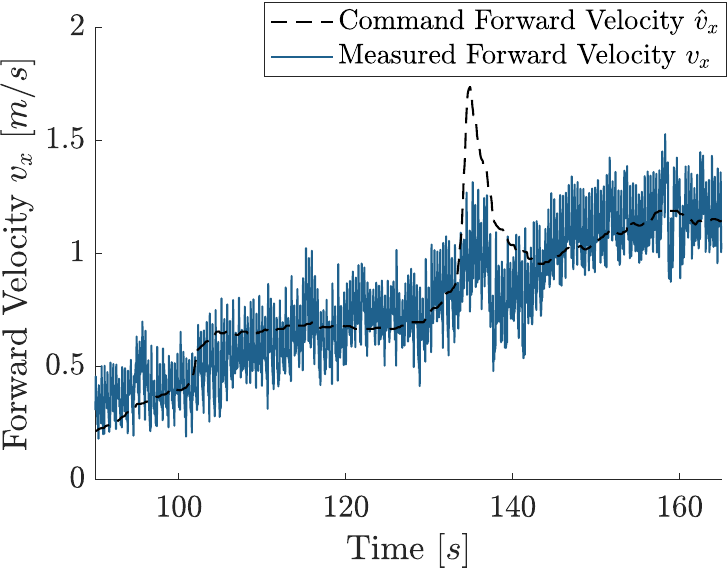}
    \caption{Forward Walking}
    \label{fig:forward_walking_plot}
  \end{subfigure}
  \vskip 5pt 
  \begin{subfigure}[b]{0.8\linewidth}
    \includegraphics[width=\linewidth]{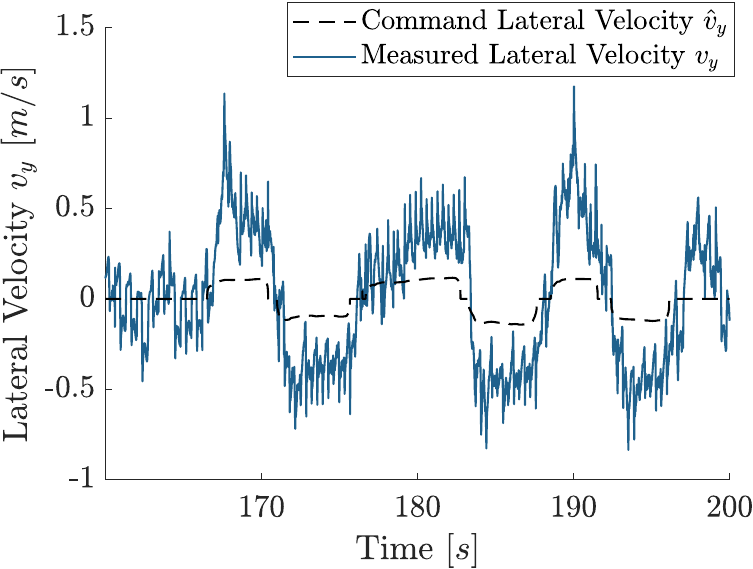}
    \caption{Turning}
    \label{fig:turning_plot}
  \end{subfigure}
  \caption{Velocity tracking plot for both forward walking and turning. Fig.~\ref{fig:forward_walking_plot} corresponds to Fig.~\ref{fig:forward_walking}, and Fig.~\ref{fig:turning_plot} corresponds to Fig.~\ref{fig:turning}. Despite the presence of noise in the base linear velocity readings from the state estimator, our policy is able to track the velocity command and execute the given locomotion tasks.}
  \label{fig:forward_turning_plot}
\end{figure}

\textit{1) Forward walking}: We evaluated the robot's ability to follow a forward velocity command on flat terrain using a treadmill, as shown in Fig.~\ref{fig:forward_walking}, confirming its capacity to walk at speeds up to 1.5 m/s. Notably, the robot demonstrated a heel-to-toe motion closely resembling human walking. Despite the noise in the base linear velocity from the state estimator, the policy enabled stable walking while accurately tracking velocity commands, as shown in Fig.~\ref{fig:forward_walking_plot}.

\textit{2) Dynamic turning}: Dynamic locomotion tasks including 90-degree and 180-degree turns were evaluated, with the results showcased in the supplementary video. Due to spatial limitations of the testing area, only small lateral velocity commands could be issued, resulting in the robot's inability to track these commands precisely. However, the robot was still able to execute stable turns as demonstrated in Fig.~\ref{fig:turning}, and Fig.~\ref{fig:turning_plot}.

\section{CONCLUSION AND FUTURE WORKS }\label{sec:conclusion}

In this work, we present an approach that combines LIPM with RL to learn the policy capable of accurately tracking desired step locations determined by LIP dynamics. Specifically, our control framework forward predicts the robot states and determines the desired step location to track a given velocity command based on LIP dynamics. We demonstrated our approach on MIT Humanoid and confirmed that tracking these steps enables stable forward walking and dynamic turning. The learned policy further showcased flexibility and adaptability by adjusting desired steps during the swing phase proving its extendability to unseen and uneven terrains. We were able to deploy our policy on MIT Humanoid achieving a forward walking speed of 1.5 m/s and dynamic 90 and 180-degree turning.

In future work, our aim is twofold: 1) We plan to incorporate vision algorithms into our system to detect the height of the terrain. This will allow us to identify stable stepping locations, enhancing the robot's ability to navigate real-world uneven terrain. 2) We aim to refine our method of determining desired step locations by replacing the LIP dynamics with whole-body dynamics, employing a model predictive controller. This refinement is expected to further improve our control framework to predict better step locations across various locomotion tasks.

\section*{ACKNOWLEDGMENT}
We thank the members of the Biomimetic Robotics Laboratory at MIT for insightful discussions and feedback on the paper.
We especially thank Se Hwan Jeon, Elijah Stanger-Jones, and Charles Khazoom for their helpful support in setting up and conducting the hardware experiments.
\printbibliography

\end{document}